\documentclass[12pt]{article}

\usepackage{sbc-template}
\usepackage{graphicx,url}

\usepackage{multirow}
\usepackage{multicol}
\usepackage{color, colortbl}
\usepackage{amsfonts}
\usepackage{mathtools}

\usepackage[latin1]{inputenc}  

\definecolor{TableHeader}{rgb}{0.7,0.7,0.7}
\definecolor{TableHeader2}{rgb}{0.5,0.5,0.5}
     
\title{Hierarchical Classification of Financial Transactions Through Context-Fusion of Transformer-based Embeddings and Taxonomy-aware Attention Layer}

\author{Antonio J. G. Busson\inst{1,2}, Rafael Rocha\inst{1,2}, Rennan Gaio\inst{1}, Rafael Miceli\inst{1}, \\  Ivan Pereira\inst{2}, Daniel de S. Moraes\inst{2}, Sérgio Colcher\inst{2}, Alvaro Veiga\inst{2}, \\Bruno Rizzi\inst{1}, Francisco Evangelista\inst{1}, Leandro Santos\inst{1}, Fellipe Marques\inst{1}, \\Marcos Rabaioli\inst{1}, Diego Feldberg\inst{1}, Debora Mattos\inst{1}, João Pasqua\inst{1}, Diogo Dias\inst{1}} 

\address{BTG Pactual \\
São Paulo -- SP -- Brazil
\nextinstitute
  Pontifical Catholic University of Rio de Janeiro (PUC-Rio) \\
  Rio de Janeiro -- RJ -- Brazil
\email{\{antonio.busson, rafael-h.rocha, rennan.gaio\}@btgpactual.com}
}

\begin{document} 

\maketitle

\begin{abstract}
This work proposes the Two-headed DragoNet, a Transformer-based model for hierarchical multi-label classification of financial transactions. Our model is based on a stack of Transformers encoder layers that generates contextual embeddings from two short textual descriptors (merchant name and business activity), followed by a Context Fusion layer and two output heads that classify transactions according to a hierarchical two-level taxonomy (macro and micro categories). Finally, our proposed Taxonomy-aware Attention Layer corrects predictions that break categorical hierarchy rules defined in the given taxonomy. Our proposal outperforms classical machine learning methods in experiments of macro-category classification by achieving an F1-score of 93\% on a card dataset and 95\% on a current account dataset.
\end{abstract}


\section{Introduction}
\label{sec:intro}

Artificial intelligence technologies, such as Machine Learning (ML), are already transforming several industries and services, including banking. Among several financial applications, ML-based technologies can accurately classify financial transactions, which can be used to drive business decisions and understand customers' financial needs \cite{belanche2019artificial}.

Current industry standards for identifying business types of merchants, such as ISO-18245\footnote{\url{https://www.iso.org/standard/33365.html}}, could be used to classify transactions via mapping rules. However, this approach has limitations: (1) old standards do not cover many actual business activities, e.g., different internet/app-based merchants are classified generically as digital products; (2) in some cases, merchants are registered with inappropriate business categories in payment systems. In this context, we hypothesize that ML-based models can overcome these limitations by learning contextual representations from a given merchant name and its related business activity and using it to classify transactions correctly, e.g., given a merchant named "John's Barbecue", contextually, the model should classify it as a restaurant, even if its registered business activity is a pharmacy related description.

Our proposed transaction classifier is based on the following:
(1) A stack of Transformer encoders~\cite{vaswani2017attention} to generate contextual embeddings from the transaction's merchant name and business activity descriptor; (2) A context-fusion layer to aggregate contextual embeddings; (3) Two output heads and a taxonomy-aware attention layer to predict hierarchical dual-labels based in a given categorical taxonomy. To demonstrate our proposal's effectiveness, we conducted benchmark experiments on two real-world datasets,  the first composed of 151,867 card transactions and the second composed of 151.838 current account transactions.

The remainder of this paper is structured as follows. We begin, in Section 2, by presenting recent related works that focus on the classification of financial transactions. In Section 3, we present our datasets. Next, Section 4 introduces our proposal, followed by Section 5, where we describe the experiments. Finally, Section 6 is devoted to our final remarks and conclusions.

\section{Related Work}

Recent works have been devoted to financial transaction classification using ML-based methods. Many focus on problems related to fraud and AML (Anti-Money Laundering). \cite{khrestina2017development} proposed a system to classify suspicious transactions. They modeled the user behavior with a relationship graph and used logistic regression to detect fraudulent activities. \cite{de2018customized} proposed the Fraud-BCN, a Bayesian network for classifying risky transactions. Their approach performed better than classic algorithms such as logistic regression, random forest, and SVM. \cite{fiore2019using} investigated the applicability of generative models as a data augmentation technique for a transactional fraud dataset. Experiments show that a classifier trained on the augmented set outperforms the same classifier trained on the original data, resulting in an effective fraud detection mechanism.

Similar to ours, other recent works also use Transformer-based methods for generating contextual embeddings of financial data. \cite{padhi2021tabular, hewapathirana2022systematic}, for instance, explored Transformer-based models as an extractor of embeddings in tabular data to detect fraudulent transactions. Experiments show that the embeddings generated by this method can improve the results of other supervised and non-supervised models,  such as Gradient Boosting and K-means. Unlike these works, we opted for a more straightforward approach since our proposal is based on the fusion of contextual embeddings extracted only from two transaction information, the merchant's name and its economic activity.

Other works focus on transaction classification for consumer behavior. \cite{cheng2022classification} proposed a method to classify individuals' spending behavior based on their monthly transaction records. They propose a classification model using k-means clustering and a neural network to categorize spending behavior based on income and spending. Their results indicate that the proposed method can classify spending behaviors such as ``low-income and high-spending" or ``high-income and low-spending". \cite{garcia2020identifying} proposed a PFM app called CoinScrap. Their solution for transaction classification is based on a two-stage method that combines a short text similarity detector with an SVM classifier.

\cite{chinchia2020} employed neural networks to categorize merchant business types using: (1) temporal features of the merchant transaction history and (2) the affinity relationship between merchants. Like ours, they perform feature-fusion between these two feature types before feeding them to a classification layer. However, their raw affinity vector is the size of the number of merchants being classified, which may occur in poor scalability. They address this problem by limiting the number of active relationships to a parameter \emph{k} that tunes the trade-off between performance and accuracy.

The hierarchical classification of financial transactions was investigated by \cite{vollset2017making}. They explored the use of external semantic data from the Brønnøysund Registry and the Google Places API to improve the accuracy of their bank transaction classification system. Similar to our datasets, their dataset consists of unstructured transaction descriptions, each labeled with a corresponding category and sub-category from 10 categories and 63 sub-categories. They used a Bag-of-words representation and a Logistic Regression as a classification method. In Experiments, their results show improvement from using enhancement of both bases separately, and a better result, when combining information of the two bases. Rather than multiple models for multi-label classification, our approach consists of a single end-to-end model trained to classify hierarchical multi-labels.

\section{Dataset}


We built two datasets for transaction classification in a retail banking context. Transactions were extracted from BTG Pactual Banking's transactional stream with a 3-month window, April - June of 2021. No information about banking customers is present in the dataset's transactions. Each transaction contains the following data: (1) a merchant name; And (2) a description of the business activity.

The first dataset is composed of 151.867 unique card transactions. In this dataset, the merchant's activity is described by the Merchant Category Code (MCC), a textual description of 296 retail financial services listed in ISO-18245. The second dataset consists of 151.838 unique account transactions. Instead of MCC, the merchant's activity of this dataset is described by the Brazilian National Classification of Economic Activities (CNAE)\footnote{\url{https://concla.ibge.gov.br/estrutura/atividades-economicas-estrutura/cnae}}, a textual description in pt-br of 695 economic activities.

Table 1 shows the taxonomy of retail services defined by BTG Pactual's consumer banking specialists. The taxonomy is structured into two hierarchical levels, the top-level entities, called Macro categories, describe broad retail sectors, e.g., Food, Shopping, and Personal Care. Each Macro category contains subcategories, called Micro categories, which describe specific retail sectors, e.g., the Shopping category contains the subcategories Electronics, Toys, and Sporting Goods. The taxonomy contains 82 micro categories, partitioned into 15 macro categories.

\begin{table*}[h]
    \centering
    \caption{Taxonomy of Retail Services}
    \scalebox{0.62}{
    \begin{tabular}{|lll|lll|lll|}
        \hline
         \rowcolor{TableHeader}\textbf{\#} & \textbf{Macro Cat.} & \textbf{Micro Cat.} & \textbf{\#} & \textbf{Macro Cat.} & \textbf{Micro Cat.} & \textbf{\#} & \textbf{Macro Cat.} & \textbf{Micro Cat.} \\
        \hline
        1 & Food & Restaurant & 29 & Health & Doctor & 57 & Bill & Water  \\
        2 & Food & Bar & 30 & Health & Health Insurance & 58 & Bill & Condominium  \\
        3 & Food & Cafe \& Bakery & 31 & Health & Therapy & 59 & Bill & Rental  \\
        4 & Food & Other Food & 32 & Health & Dentist & 60 & Bill & Energy  \\
        5 & Groceries & Horticulture & 33 & Health & Pharmacy & 61 & Bill & Gas \\
        6 & Groceries & Beverages & 34 & Health & Exams & 62 & Bill & Telecomunication \\
        7 & Groceries & Butchery & 35 & Health & Other Health & 63 & Bill & Credit Card  \\
        8 & Groceries & Supermarket & 36 & Personal Care & Beauty Salon & 64 & Bill & Other Bill  \\
        9 & Groceries & Other Groceries & 37 & Personal Care & Spa \& Esthetic Clinic & 65 & Entertainment & Clubs  \\
        10 & Shopping & Clothing \& Accessories & 38 & Personal Care & Sports & 66 & Entertainment & Games  \\
        11 & Shopping & Books \& Supplies & 39 & Personal Care & Gym & 67 & Entertainment & Shows \& Events  \\
        12 & Shopping & Eletronics & 40 & Personal Care & Cosmetics & 68 & Entertainment & Cinema \& Theater  \\
        13 & Shopping & Toy & 41 & Personal Care & Tattoo \& Piercing & 69 & Entertainment & Expositions  \\
        14 & Shopping & Sporting Goods & 42 & Personal Care & Other Personal Care & 70 & Entertainment & Streaming  \\
        15 & Shopping & Gifts & 43 & Pets & Pet Shop & 71 & Entertainment & Journals \& Magazines  \\
        16 & Shopping & Other Shopping & 44 & Pets & Veterinary & 72 & Entertainment & Parks  \\
        17 & Transport & Fuel & 45 & Pets & Other Pets & 73 & Entertainment & Other Entertainment  \\
        18 & Transport & Car Maintenace & 46 & Home & Home Goods & 74 & Education & Courses  \\
        19 & Transport & Bike & 47 & Home & Renovations & 75 & Education & School  \\
        20 & Transport & Parking & 48 & Home & Gardening \& Pool & 76 & Education & College  \\
        21 & Transport & Car Wash & 49 & Home & Laundry & 77 & Education & Other Education  \\
        22 & Transport & Public Transportation & 50 & Home & Domestic Employees & 78 & Travel & Accommodation  \\
        23 & Transport & Taxi \& Apps & 51 & Home & Security & 79 & Travel & Tickets  \\
        24 & Transport & Car Rental & 52 & Home & Other Home & 80 & Travel & Exchange  \\
        25 & Transport & Traffic Ticket & 53 & Tax \& Tribute & House Taxes & 81 & Travel & Other Travel  \\
        26 & Transport & Toll & 54 & Tax \& Tribute & Car Taxes & 82 & Donation & Other Donation  \\
        27 & Transport & Other Transport & 55 & Tax \& Tribute & Income Tax & &  &   \\
        28 & Other Category & Other & 56 & Tax \& Tribute & Other Tax \& Tribute & &  &   \\
        \hline
    \end{tabular}
    }
    \label{tab:microcategories}
\end{table*}

A group of 8 volunteers manually annotated the dataset in two steps. In the first step, we partition our two datasets into four equal parts. Next, we formed four pairs of volunteers to each label a specific part of the dataset. In the second step, we collect all transactions with labeling disagreement between the pairs and create a new subset of transactions. Finally, this subset was divided equally among the volunteers, where each one voted for one of the labels suggested in the previous step. Table 2 shows the distribution of transactions by macro categories in our two datasets.

\begin{table}[h]
    \centering
    \caption{Dataset distribution by Macro category}
    \scalebox{0.7}{
    \begin{tabular}{|rlrr|}
        \hline
         \rowcolor{TableHeader}\textbf{\#} & \textbf{Macro Categ.} & \textbf{Card} & \textbf{Current Acc.} \\
       \hline
        1 & Shopping  & 40,194 & 37,369 \\
        2 & Food  & 30,405 & 28,124 \\
        3 & Groceries  & 21,559 & 20,888 \\
        4 & Home  & 19,640 & 21,411 \\
        5 & Transport  & 18,175 & 18,588 \\
        6 & Health  & 10,426 & 10,599 \\
        7 & Personal Care  & 10,272 & 10,186 \\
        8 & Bill  & 8,306 & 10,230 \\
        9 & Education & 4,514 & 5,085 \\
        10 & Pets  & 3,645 & 3,512 \\
        11 & Travel  & 2,116 & 2,298 \\
        12 & Entertainment  & 1,962 & 2,033 \\
        13 & Donation  & 1,745 & 2,137 \\
        14 & Other Category  & 1,098 & 841 \\
        15 & Tax and Tribute  & 70 & 58 \\
        \hline
    \end{tabular}
    }
    \label{tab:category_distribution}
\end{table}



\section{Model}
\label{sec:plataform_overview}

The overview of our proposed method is depicted in Figure 1. The Two-headed DragoNet comprises a stack of Transformer layers that generates contextual embeddings for two textual inputs. Next, a Context Fusion layer aggregates these contextual embeddings to generate a single enhanced contextual representation. Finally, two output heads classify the transaction according to the hierarchy defined by a given taxonomy; the first head predicts top-level classes (macro categories), while the second head predicts second-order classes (micro categories). The Taxonomy Attention layer uses the first head output to correct the hierarchy inconsistencies of the second head output.

\begin{figure*}[!ht]
\centering
\includegraphics[width=\textwidth]{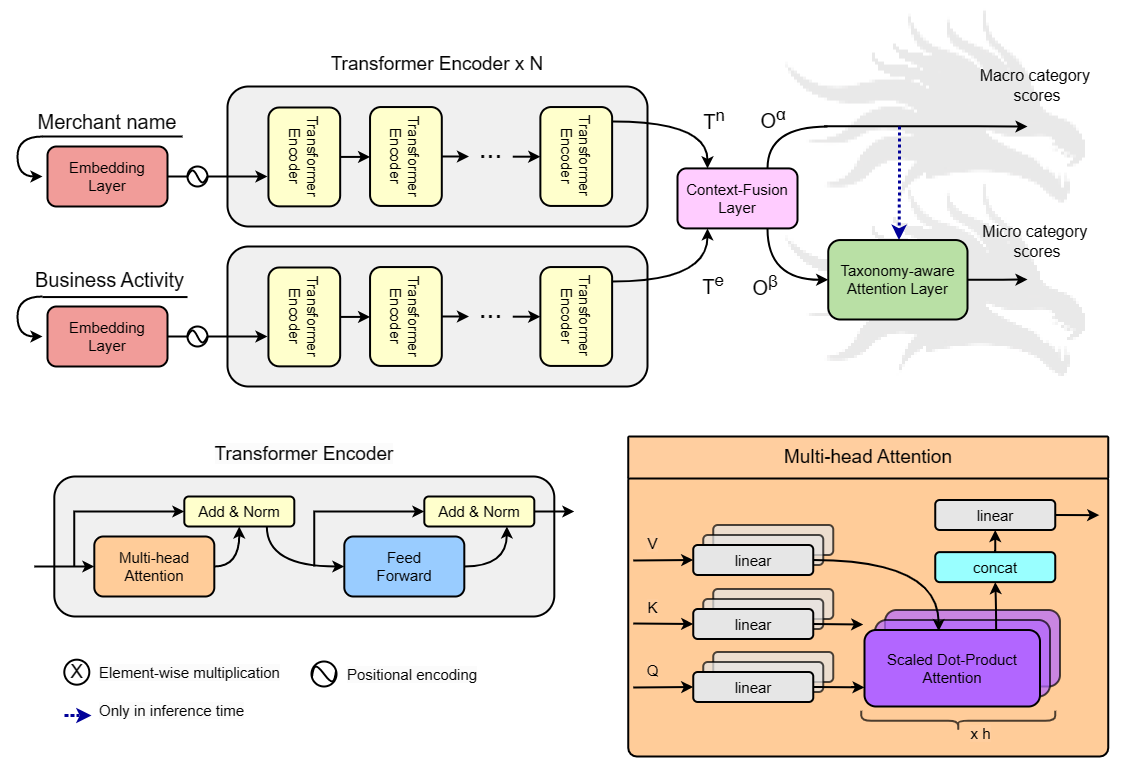}
\caption{Two-headed DragoNet architecture.}
\label{fig:dragonet_architecture}
\end{figure*}

The subsections that follow are structured as follows. In Section 4.1, we detail the Two-head DragoNet model workflow. Next, in Section 4.2, we describe the Transformer Encoder layer. In Section 4.3, we introduce the Context-Fusion layer. Finally, in Section 4.4, we detail the Taxonomy-aware Attention layer.

\subsection{Model workflow}

Given a collection of financial transactions $B$, each transaction $b \in B$ is defined as a 2-tuple $b\equiv  \langle x,y \rangle$, where $x$ and $y$ denotes a transactional feature and a category, respectively. Each feature is defined as $x \equiv \langle x^{n}, x^{e} \rangle$, where $x^{n}$ is a merchant name, and $x^{e}$ is a business activity description. Both $x^{n}$ and $x^{e}$ comprise an array $\{x_1, x_2, ..., x_m~|~x_i \in S \}$,  where $m$ is the number of words in the sentence, and $S$ is the set of word tokens of the dataset's text corpus. Given a categorical taxonomy as a DAG $G \equiv \langle C, R \rangle$, in which $C$ is a set of categories, and $R \subseteq C \times C$ is a set of relationships. Each transactional category consists of a dual-label $y \equiv \langle y_{i}, y_{j} \rangle \in R ~|~ y_i \downarrow y_j$, where $\downarrow$ denotes a hierarchy relationship, i.e., $y_i$ is the parent of $y_j$.

Using the column embedding layer ($E_\theta$), we generate parametric embeddings for both $x^{n}$ and $x^{e}$. Let $E_\theta(x) \equiv \{ e_{\theta_1(x_1)}, e_{\theta_2(x_2)}, ..., e_{\theta_m(x_{m})}\}$ where $e_{\theta_i(x_i)} \in \mathbb{R}^d$ is the parametric embedding of the i-th sentence word. Next, we add ``positional encodings"~($P$) to these parametric embeddings $E_\theta(x^n)$ and $E_\theta(x^e)$ to inject sequential information, since the vanilla Transformer contains no recurrence and no convolution. 

Afterward, each modality of positional-encoded embeddings is inputted into a stack of transformer encoders. Then, each positional-encoded embedding is transformed into contextual embedding when outputted from the last Transformer Encoder layer through the consecutive aggregation of context from embeddings generated by inner Transformer Encoder layers. Let $T_\phi$ be a function that represents the stack of Transformer encoder layers. $T_\phi$ operates the positional encoded embeddings ($ E_\theta(x) + P$) and returns the correspondent contextual embeddings $T \equiv \{ t_1, t_2, .., t_n\}$, where $t_i \in \mathbb{R}^d for~i \in \{1, ..., n\}$.

Let $T^n$ and $T^e$ be the contextual embeddings generated from $x^n$ and $x^e$ using $T_\phi$. We denote the Context Fusion layer as a function $F_\lambda$, in which it aggregates both $T^n$ and $T^e$ to produce a single high-level contextual representation. The output of this layer is defined by the parameter $\lambda \in \{\alpha,\beta\}$, which returns $O^\alpha = \{o^\alpha_1, o^\alpha_2, ..., o^\alpha_j\}$ if $\lambda=\alpha$ or $O^\beta = \{o^\beta_1, o^\beta_2, ..., o^\beta_k\}$ if  $\lambda=\beta$, where $O^\alpha$ and $O^\beta$ are the prediction distribution for all macro and micro categories, respectively.

Let $H$ be the categorical cross-entropy function. We minimize the following multi-task loss function: 
$$
\mathcal{L}( \langle x^n, x^e \rangle,  \langle y^\alpha, y^\beta \rangle ) = L^\alpha + L^\beta
$$
\vspace{-15px}
$$
\mathcal{L}^\alpha = H(F_\alpha(T_\phi ( E_\theta(x^n)+P, E_\theta(x^e))+P, y^\alpha) 
$$
\vspace{-15px}
$$
\mathcal{L}^\beta = H(F_\beta(T_\phi ( E_\theta(x^n)+P, E_\theta(x^e))+P, y^\beta) 
$$
where $\mathcal{L}^\alpha$ and $\mathcal{L}^\beta$ are loss functions for macro and micro categories classification, respectively.

Exclusively at inference time, we use the Taxonomy-aware Attention layer to adjust predictions of micro categories ($O^\beta$) that break the hierarchy relationships defined in the given taxonomy $G$, producing an $O'^\beta $ output. 

\subsection{Transformer Encoder Layer}

The Transformer Encoder layer learns to generate contextual representations (embeddings) from a sequence of symbolic inputs using stacked Self-Attention and Fully Connected layers. \textit{Attention} is a function that maps a Query ($Q$) and a set of Key ($K$) and Value ($V$) pairs to an output. Formally, let $Q \in \mathbb{R}^{m \times k}$, $K \in \mathbb{R}^{m \times k}$, and $V \in \mathbb{R}^{m \times v}$, where $m$ is the number of the embeddings inputted; And $k$ and $v$ are the dimensions of the Key and Value vectors, respectively. The Scaled Dot-Product Attention block computes the dot products of the query with all keys, divides each by $\sqrt{k}$, and applies a softmax function to obtain the weights on the values. The Attention function is defined as follows:
$$
Attention(Q,K,V) = A.V, \\
$$
where $A = softmax( (QK^T)/ \sqrt{k})$. For each embedding, the attention matrix $A \in \mathbb{R}^{m \times m}$ calculates how much it attends to other embeddings, thus transforming the embedding into a contextual one.  

Instead of a single attention function. In parallel,  Multi-head Attention linearly projects the queries, keys, and values $h$ times with different, learned linear projections. Then, these are concatenated and once again projected as follows: 
$$
MultiHead(Q,K,V) = Concat(head_1, head_2, ..., head_h)W^O,
$$
where $head_i = Attention(QW_i^Q, KW_i^K, VW_i^V)$, projections are parameter matrices $W_i^Q \in \mathbb{R}^{m \times k}$, $W_i^K \in \mathbb{R}^{m \times k}$, $W_i^V \in \mathbb{R}^{m \times v}$, and $W_i^O \in \mathbb{R}^{hv \times m}$.

In addition to Transformer Encoder sub-layers, the output of the Multi-head Attention is projected back to the embedding of dimension $d$ through a fully connected feed-forward network (FFN) composed of two linear transformations with a ReLU activation.

\subsection{Context-Fusion Layer}
The Context-Fusion Layer comprises a concatenation of the contextual embeddings from merchant name and business activity description ($T_{n}$ and $T_{e}$), an FNN with ReLU activation, followed by a last layer of a linear transformation with Sigmoid activation:
$$
ContextFusion(T^n, T^e, \lambda) = Sigmoid^\lambda(FNN(Concat(T^{n}, T^{e})))
$$
where the parameter $\lambda \in \{\alpha,\beta\}$ denotes the size in units of the sigmoidal output, if $\lambda=\alpha$, the output size is the number of macro categories, else if $\lambda=\beta$ the size is the number of micro categories.

\subsection{Taxonomy-aware Attention Layer}
The Taxonomy-aware Attention layer consists of a rule-based layer that adjusts predictions of micro categories that break the category hierarchy defined in a taxonomy. Given $O^\alpha$ and $O^\beta$, this layer produces an array $M$ of size $b$. 
We assume that each position of $M$ corresponds to a micro category. Then, positions corresponding to subcategories of the macro categories with the highest score are filled with 1 and 0 otherwise. Finally, a point-wise multiplication between $M$ and $O^\beta$ suppresses all invalid category scores:
$$
TaxonomyAttention(O^{\alpha}, O^{\beta}) = M.O^{\beta}
$$
where:
$$
M = \{m_1, m_2, ..., m_{b}\}~|~m_i \in \{0,1\}~for~i \in \{1,...b \}
$$
\[
    m_i = 
\begin{dcases}
    1, & \text{if } Class(O^{\beta}_i) \downarrow Class(Argmax(O^{\alpha})) \\
    0, & \text{otherwise}
\end{dcases}
\]

\section{Experiments}

This section evaluates our models' effectiveness in classifying financial transactions. We perform benchmark comparisons using different algorithms: the nearest neighbor heuristic (KNN), shallow classifiers like SVC (C-Support Vector Classification) and Random Forest, and other deep learning methods such as GRU (Gated Recurrent Units), LSTM (Long Short-Term Memory), and BLSTM (Bidirectional LSTM). 

We encode dataset texts using three different techniques: TF-IDF, HashingVectorizer, and Tokenizer (Index-Based Encoding). We used the Tokenizer technique to feed the deep learning models' parametric embedding layer. In contrast, the other two techniques were combined with the shallow models since they are more suitable for these algorithms.

We perform three experiments for both datasets to understand how each data type contributes to the classification task. The first experiment uses just the merchant name data, the second uses just the business activity description data, and the last combines both data. For each experiment, we perform the same 10-fold split and evaluate the models by the Precision (P), Recall (R), and F1-Score (F1).

\subsection{Results}

Table 3 shows the result of the experiment using only the merchant name data. Notably, models based on deep learning produced superior results than other baseline models. The Transformer model achieved the best result, producing an F1-score of 56.78\% in the card dataset and 59.14\% in the current account dataset for macro-category classification. Regarding micro-category classification, it produced an F1-score of 46.02\% in the card dataset and 45.15\% in the current account dataset.

In this scenario, we noticed that more than the merchant name alone is needed for the classification task, as our models produced a low F1 score. Figure 2 (A) shows the contextual embeddings of the merchant names generated by the transformer model. These embeddings have a strong interclass correlation, making the classification task challenging. When inspecting the merchant names, we noticed that it is impossible to determine the categories with confidence in many cases, as some merchant names are, for example, person names or generic names without precise semantics. e.g., Red Shop, Thunder, Bluebird.




\begin{table}[h]
    \caption{Experiment results in scenario 1: Using only the merchant name data}
      \centering
        \scalebox{0.8}{
        \begin{tabular}{|l|ccc|ccc|}
        \hline
        \rowcolor{TableHeader2} \multicolumn{7}{|c|}{\textbf{(A) Card Transactions Dataset }} \\
        \hline
          \rowcolor{TableHeader}  &    \multicolumn{3}{c|}{\textbf{Macro Category}} & \multicolumn{3}{c|}{\textbf{Micro Category}} \\
          \rowcolor{TableHeader}  \textbf{Model} & \textbf{P} & \textbf{R} & \textbf{F1} & \textbf{P} & \textbf{R} & \textbf{F1} \\
        \hline
        Transformer & 57.76 & 55.85 & 56.78 & 47.86 & 44.33 & 46.02 \\
        BLSTM & 58.86 & 53.90 & 56.27 & 47.72 & 40.29 & 43.69 \\
        LSTM & 59.47 & 53.36 & 56.24 & 49.98 & 40.98 & 45.03 \\
        GRU & 58.85 & 53.28 & 55.92 & 47.77 & 40.90 & 44.06 \\
         TF-IDF + SVC & 45.77 & 34.66 & 39.44 & 24.33 & 20.01 & 21.95 \\ 
         TF-IDF + RF & 40.46 & 36.89 & 38.59 & 28.25 & 23.31 & 25.54 \\ 
         HashingVec. + RF & 31.16 & 31.91 & 31.53 & 19.49 & 19.27 & 19.37 \\ 
         HashingVec. + SVC & 30.19 & 31.19 & 30.68 & 12.19 & 16.00 & 13.83 \\
         TF-IDF + KNN & 37.51 & 23.00 & 28.51 & 21.39 & 16.46 & 18.60 \\ 
         HashingVec. + KNN & 25.01 & 23.97 & 24.47 & 14.79 & 16.22 & 15.47 \\ 
        \hline
        \rowcolor{TableHeader2} \multicolumn{7}{|c|}{\textbf{(B) Current Account Transactions Dataset }} \\
        \hline
          \rowcolor{TableHeader}  & \multicolumn{3}{c|}{\textbf{Macro Category}} & \multicolumn{3}{c|}{\textbf{Micro Category}} \\
          \rowcolor{TableHeader} \textbf{Model} & \textbf{P} & \textbf{R} & \textbf{F1} & \textbf{P} & \textbf{R} & \textbf{F1} \\
        \hline
         Transformer & 61.37 & 57.08 & 59.14 & 47.62 & 42.93 & 45.15 \\
         BLSTM & 61.85 & 56.60 & 59.10 & 49.77 & 43.94 & 46.67 \\
         GRU & 60.43 & 57.16 & 58.74 & 50.85 & 43.84 & 47.08 \\
         LSTM & 59.38 & 56.06 & 57.67 & 46.17 & 42.64 & 44.33 \\
        TF-IDF + Random Forest & 39.67 & 28.34 & 33.06 & 15.91 & 16.42 & 16.16 \\
        HashingVec. + Random Forest & 28.30 & 23.47 & 25.65 & 18.65 & 14.78 & 16.49 \\
        HashingVec. + SVC & 32.87 & 19.56 & 24.52 & 09.90 & 09.92 & 09.90 \\
        TF-IDF + KNN & 33.08 & 19.43 & 24.48 & 19.32 & 07.92 & 11.23 \\
        TF-IDF + SVC & 27.43 & 20.18 & 23.25 & 10.87 & 09.94 & 10.38 \\   
        HashingVec. + KNN & 24.10 & 20.73 & 22.28 & 11.66 & 11.41 & 11.53 \\
         
        \hline
    \end{tabular}
    }
\end{table}

\begin{figure}[h]
\centering
\includegraphics[width=0.75\textwidth]{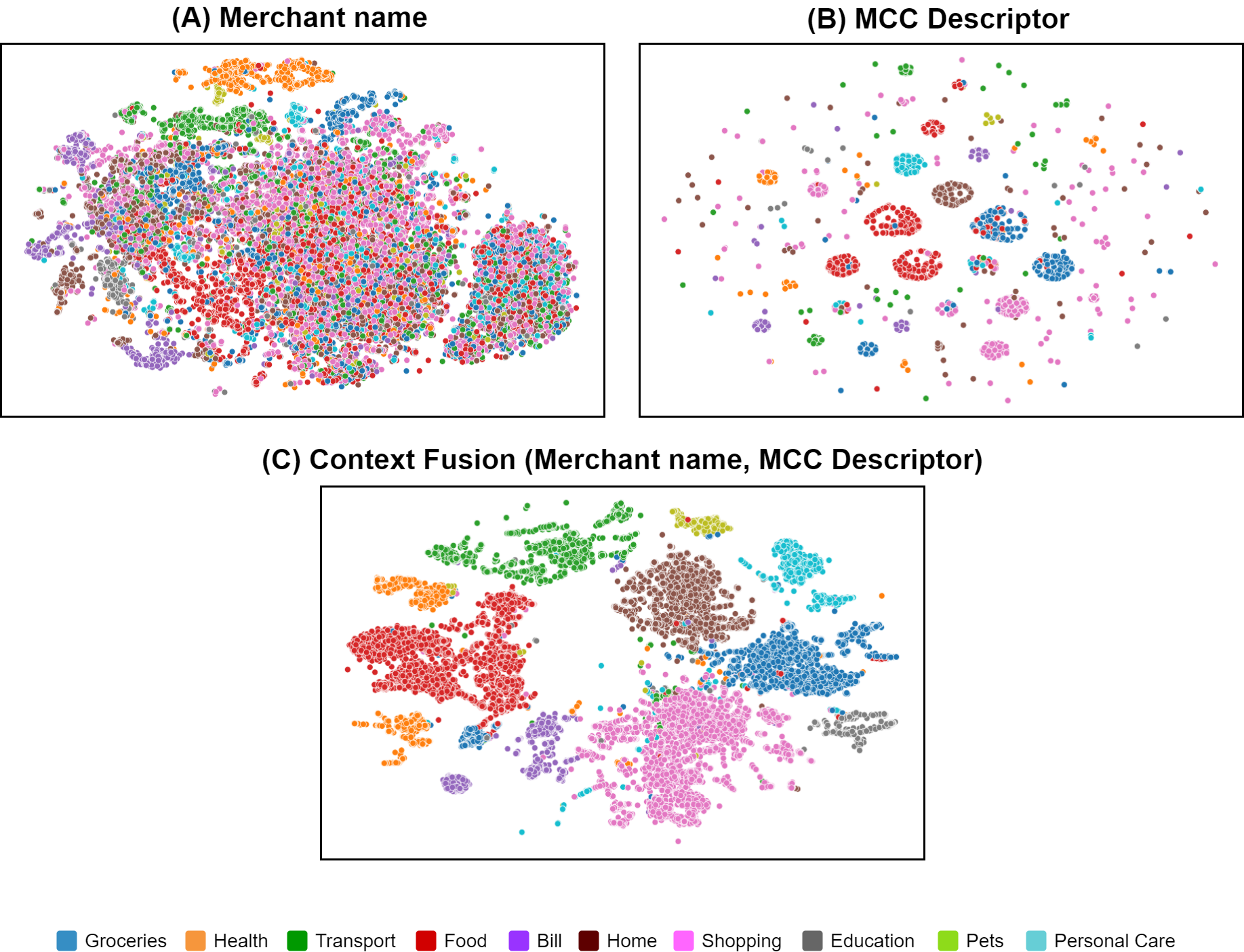}
\caption{T-SNE visualizations of Contextual Embeddings in 2D space for transactions of the top-10 macro categories.}
\label{fig:cnn_inception_resnet_mod}
\end{figure}

Table 4 shows the experiment result using only the business activity description data. Again, the Transformer model achieved the best result, producing an F1-score of 89.07\% in the card dataset and 93.24\% in the current account dataset for macro-category classification. Regarding micro-category classification, the Transformer model produced an F1-score of 76.98\% in the card dataset and 85.41\% in the current account dataset. When comparing with the first experiment, we noticed that using the description of the business activity instead of the name produced an F1-score gain of more than 30\% with both datasets. 

\begin{table}[h]
    \caption{Experiment results in scenario 2: Using only the business activity descritor}
      \centering
        \scalebox{0.8}{
\begin{tabular}{|l|ccc|ccc|}
        \hline
        \rowcolor{TableHeader2} \multicolumn{7}{|c|}{\textbf{(A) Card Transactions Dataset }} \\
        \hline
          \rowcolor{TableHeader}  & \multicolumn{3}{c|}{\textbf{Macro Category}} & \multicolumn{3}{c|}{\textbf{Micro Category}} \\
          \rowcolor{TableHeader} \textbf{Model} & \textbf{P} & \textbf{R} & \textbf{F1} & \textbf{P} & \textbf{R} & \textbf{F1} \\
        \hline
         Transformer & 89.52 & 88.64 & 89.07 & 76.34 & 77.63 & 76.98 \\
         BLSTM & 86.02 & 84.47 & 85.23 & 69.08 & 69.01 & 69.04 \\
         GRU & 81.93 & 80.52 & 81.21 & 67.96 & 69.36 & 68.65 \\
         LSTM & 81.96 & 80.47 & 81.20 & 68.23 & 69.03 & 68.62 \\
         
         HashingVec. + Random Forest & 82.52 & 81.72 & 82.11 & 66.50 & 70.09 & 68.24 \\
         HashingVec. + SVC & 81.40 & 80.49 & 80.94 & 63.41 & 67.80 & 65.53 \\
         HashingVec. + KNN & 76.35 & 74.14 & 75.22 & 59.77 & 64.28 & 61.94 \\
         TF-IDF + Random Forest & 78.23 & 72.16 & 75.07 & 60.28 & 61.55 & 60.90 \\
         TF-IDF + SVC & 75.92 & 71.86 & 73.83 & 51.78 & 58.61 & 54.98 \\ 
         TF-IDF + KNN & 73.77 & 64.91 & 69.05 & 58.77 & 56.41 & 57.56 \\

        \hline
        \rowcolor{TableHeader2} \multicolumn{7}{|c|}{\textbf{(B) Current Account Transactions Dataset }} \\
        \hline
          \rowcolor{TableHeader} & \multicolumn{3}{c|}{\textbf{Macro Category}} & \multicolumn{3}{c|}{\textbf{Micro Category}} \\
          \rowcolor{TableHeader} \textbf{Model} & \textbf{P} & \textbf{R} & \textbf{F1} & \textbf{P} & \textbf{R} & \textbf{F1} \\
        \hline
        Transformer & 93.57 & 92.93 & 93.24 & 86.37 & 84.49 & 85.41 \\
        GRU & 88.77 & 88.10 & 88.43 & 80.14 & 78.69 & 79.40 \\
        LSTM & 88.67 & 87.68 & 88.17 & 79.68 & 78.22 & 78.94 \\
        BLSTM & 88.19 & 87.44 & 87.81 & 79.58 & 77.56 & 78.55 \\
        TF-IDF + SVC & 79.81 & 70.76 & 75.01 & 53.56 & 53.07 & 53.31 \\  
        HashingVec. + Random Forest & 76.86 & 72.38 & 74.55 & 61.00 & 59.91 & 60.45 \\
        TF-IDF + Random Forest & 76.25 & 71.17 & 73.62 & 59.41 & 58.92 & 59.16 \\
        HashingVec. + SVC & 77.91 & 68.53 & 72.91 & 57.84 & 55.20 & 56.48 \\   
        HashingVec. + KNN & 67.98 & 62.23 & 64.97 & 42.51 & 46.04 & 44.20 \\
        TF-IDF + KNN & 62.94 & 56.02 & 59.27 & 45.81 & 42.30 & 43.98 \\
       
        \hline
    \end{tabular}

    }
\end{table}

Looking at the contextual embeddings generated from business activity descriptors in Figure 2 (B), we noticed that these embeddings formed contextual blobs, many of them with predominantly homogeneous categories. This is expected since several business activities with similar descriptors are structured in the same market groups. e.g., taxicabs, ambulance services, passenger railways, and bus lines are all activities from the transportation group. Notably, using the business activity description data is more beneficial for the classification task. However, this strategy fails when the transaction is registered with the wrong merchant activity, e.g., restaurants registered with hotel or gas station activity. 

Table 5 shows the experiment results using merchant name and business activity descriptor data. In this scenario, our proposal achieved the best result, producing an F1-score of 93.02\% in the card dataset and 95.07\% in the current account dataset for macro-category classification. Regarding micro-category classification, it achieved an F1-score of 84.54\% in the card dataset and 86.66\% in the current account dataset. Compared with second place, note that our proposed Taxonomy-aware Attention Layer (TAL) improved micro-category classification, achieving an F1-score gain of 0.9\%  on the card dataset and 1\% on the current account dataset.

\begin{table*}[h]
    \centering
    \caption{Experiment results in scenario 3: Using merchant name and business activity description}
     \scalebox{0.75}{
    \begin{tabular}{|l|ccc|ccc|}
        \hline
        \rowcolor{TableHeader2} \multicolumn{7}{|c|}{\textbf{(A) Card Transactions Dataset }} \\
        \hline
          \rowcolor{TableHeader}  & \multicolumn{3}{c|}{\textbf{Macro Category}} & \multicolumn{3}{c|}{\textbf{Micro Category}} \\
          \rowcolor{TableHeader} \textbf{Model} & \textbf{P} & \textbf{R} & \textbf{F1} & \textbf{P} & \textbf{R} & \textbf{F1} \\
        \hline
         (Two-headed DragoNet) Transformer + Context Fusion + TAL & 93.10 & 92.96 & 93.02 & 84.48 & 84.61 & 84.54 \\
          Transformer + Context Fusion & 93.10 & 92.96 & 93.02 & 83.55 & 83.78 & 83.66 \\
          Transformer  & 92.75 & 92.33 & 92.53 & 82.21 & 82.54 & 82.37 \\
          BLSTM & 92.20 & 92.04 & 92.11 & 81.73 & 82.16 & 81.94 \\
          GRU & 92.12 & 91.94 & 92.02 & 82.67 & 82.56 & 82.61 \\
          LSTM  & 91.80 & 91.63 & 91.71 & 82.09 & 81.92 & 82.00 \\
          HashingVectorizer + Random Forest & 81.45 & 81.29 & 81.36 & 65.39 & 69.14 & 67.21 \\
          TF-IDF + Random Forest & 80.67 & 81.25 & 80.95 & 65.98 & 69.08 & 67.49 \\
          HashingVectorizer + SVC & 79.12 & 77.17 & 78.13 & 59.45 & 63.34 & 61.33 \\ 
          TF-IDF + KNN & 77.90 & 76.07 & 76.97 & 62.50 & 63.34 & 62.91 \\
          HashingVectorizer + KNN & 77.14 & 76.35 & 76.74 & 61.76 & 65.73 & 63.68 \\
          TF-IDF + SVC & 77.14 & 75.32 & 76.21 & 60.14 & 61.61 & 60.86 \\

        \hline
        \rowcolor{TableHeader2} \multicolumn{7}{|c|}{\textbf{(B) Current Account Transactions Dataset }} \\
        \hline
          \rowcolor{TableHeader} & \multicolumn{3}{c|}{\textbf{Macro Category}} & \multicolumn{3}{c|}{\textbf{Micro Category}} \\
          \rowcolor{TableHeader} \textbf{Model} & \textbf{P} & \textbf{R} & \textbf{F1} & \textbf{P} & \textbf{R} & \textbf{F1} \\
        \hline
        (Two-headed DragoNet) Transformer + Context Fusion + TAL &  95.15 & 95.01 & 95.07 & 87.42 & 85.93 & 86.66 \\
         Transformer + Context Fusion & 95.15 & 95.01 & 95.07 & 86.39 & 84.97 & 85.67 \\
        Transformer & 93.76 & 93.83 & 93.79 & 85.49 & 84.71 & 85.09 \\
        BLSTM & 92.47 & 91.69 & 92.07 & 83.20 & 81.80 & 82.49 \\
        GRU & 92.19 & 91.62 & 91.90 & 84.34 & 81.73 & 83.01 \\
        LSTM & 92.17 & 91.57 & 91.86 & 84.44 & 82.16 & 83.28 \\
        
         TF-IDF + Random Forest & 75.94 & 69.10 & 72.35 & 59.07 & 56.55 & 57.78 \\
         HashingVectorizer + Random Forest & 70.80 & 62.46 & 66.36 & 54.56 & 53.35 & 53.94 \\
         TF-IDF + SVC & 70.82 & 60.64 & 65.33 & 45.48 & 44.50 & 44.98 \\
         TF-IDF + KNN & 68.53 & 60.97 & 64.52 & 49.09 & 47.11 & 48.07 \\
         HashingVectorizer + KNN & 65.83 & 59.33 & 62.41 & 44.58 & 47.82 & 46.14 \\
         HashingVectorizer + SVC & 66.52 & 56.88 & 61.32 & 42.36 & 37.37 & 39.70 \\

        \hline
    \end{tabular}
    }
    \label{tab:valid_dataset1}
\end{table*}

Notably, the strategy of fusing contextual embeddings was beneficial, as the ``Transformer + Context Fusion"~model achieved F1-score gains in both datasets compared to the vanilla transformer. Observing Figure 2 (C), note that unlike (A) and (B), context fusion generated embeddings with evident categorical clusters.

\section{Final Remarks}

In this work, we have proposed the Two-headed DragoNet, a Transformer-based model for hierarchical multi-label classification of financial transactions. Given a merchant name and its correspondent business activity description, our model generates contextual embeddings for these inputs using a stack of Transformers encoder layers. Next, a Context-Fusion layer aggregates these embeddings to generate a single high-level embedding representation. Two output heads classify transactions according to a hierarchical two-level taxonomy. Finally, our proposed Taxonomy-aware Attention Layer corrects predictions that break categorical hierarchy rules defined in the given taxonomy.

We built two datasets using actual transactions from BTG Pactual Banking's transactional stream. Our proposal outperforms classical machine learning methods in experiments by achieving a macro-category classification with an F1-score of 93\% in the card dataset and 95\% in the current account dataset. Regarding micro-category classification, our proposal achieved an F1-score of 84.5\% in the card dataset and 86.6\% in the current account dataset. Notably, the Taxonomy-aware Attention layer improved micro-category classification, achieving an F1-score gain of approximately 1\% in both datasets.

Regarding future work, we intend to investigate ways to improve the performance of our model. We plan to enrich our datasets with external semantic data, as proposed by \cite{vollset2017making}. We also intend to explore contextual customer embeddings (Customer2Vec) to help classify complex examples, especially in transactions with ambiguous merchant names. Finally, we plan to expand our taxonomy to a more extensive retail consumption ontology with unlimited hierarchy levels. To consume this data structure, we intend to enrich our model architecture with novel deep learning techniques, mainly based on Neuro-Symbolic models \cite{costa2020towards}.

\newpage


\bibliographystyle{sbc}
\bibliography{sbc-template}

\end{document}